\documentclass[10pt,twocolumn,letterpaper]{article}

\usepackage{hhline}
\usepackage{color, colortbl}
\usepackage{multirow}
\usepackage{ijcb}
\usepackage{times}
\usepackage{epsfig}
\usepackage{graphicx}
\usepackage{amsmath}
\usepackage{amssymb}
\usepackage{subcaption}
\usepackage[utf8]{inputenc}
\usepackage{lipsum} 
\usepackage{booktabs}
\usepackage{xcolor}
\usepackage{makecell}
\usepackage{url}

\usepackage{authblk}

\ijcbfinalcopy 

\ifijcbfinal\fi
\begin{document}

\title{Identity Document to Selfie Face Matching Across Adolescence}
\author[1]{Vítor Albiero}
\author[5]{Nisha Srinivas}
\author[3]{Esteban Villalobos}
\author[3]{Jorge Perez-Facuse}
\author[4]{Roberto Rosenthal}
\author[3]{Domingo Mery}
\author[2]{Karl Ricanek}
\author[1]{Kevin W. Bowyer}
\affil[1]{Department of Computer Science and Engineering, University of Notre Dame}
\affil[2]{Department Computer Science, University of North Carolina Wilmington}
\affil[3]{Department of Computer Science, Pontificia Universidad Catolica de Chile}
\affil[4]{BiometryPass, Chile}
\affil[5]{Oak Ridge National Laboratory}

\maketitle

\begin{abstract}
Matching live images (``selfies'') to images from ID documents is a problem that can arise in various applications.
A challenging instance of the problem arises when the face image on the ID document is from early adolescence and the live image is from later adolescence.
We explore this problem using a private dataset called Chilean Young Adult (CHIYA) dataset, where we match live face images taken at age 18-19 to face images on ID documents created at ages 9 to 18.
State-of-the-art deep learning face matchers (e.g., ArcFace) have relatively poor accuracy for document-to-selfie face matching.
To achieve higher accuracy, we fine-tune the best available open-source model with triplet loss for a few-shot learning.
Experiments show that our approach achieves higher accuracy than the DocFace+ model recently developed for this problem.
Our fine-tuned model was able to improve the true acceptance rate for the most difficult (largest age span) subset from $62.92\%$ to $96.67\%$ at a false acceptance rate of $0.01\%$.
Our fine-tuned model is available for use by other researchers.

\end{abstract}

\section{Introduction}

\begin{figure}[t]
    \centering
    \includegraphics[width=0.7\columnwidth]{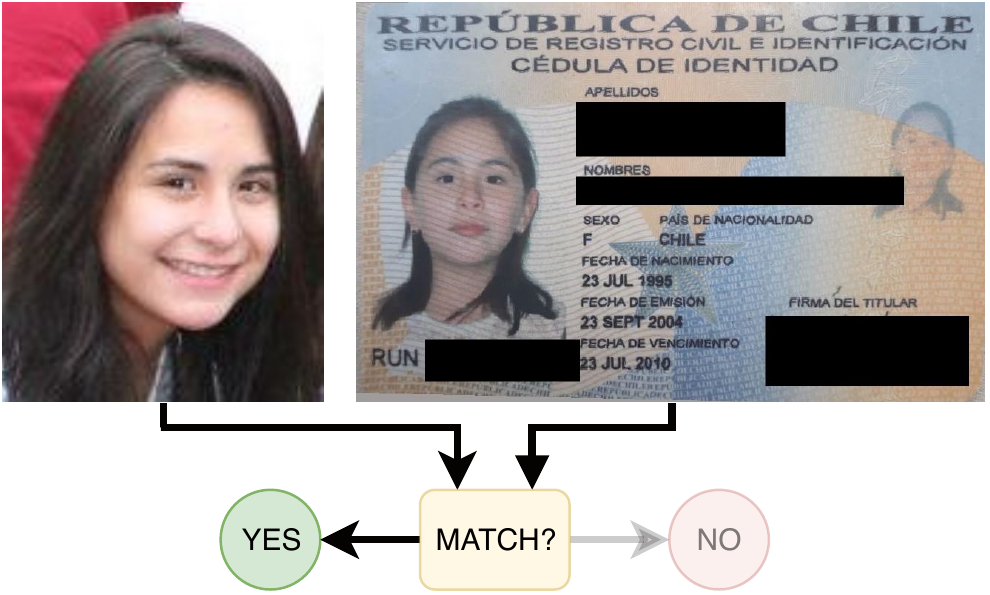}
    \caption{Matching problem: selfie (left) and ID photo (right) in this case belong to the same person. In the selfie she is 18 years old, but in the ID photo she is 9 years old.}
    \label{fig:problem}
    \vspace{-1em}
\end{figure}

The problem of matching ``live'' or ``selfie''\footnote{The term ``selfie'' has become used to describe a probe image in this context, even though the image may not be taken by the subject, meaning that it is not a literal selfie.} face images against images from an ID document (see Fig. \ref{fig:problem}) has been tackled in a number of previous works. 
Some previous works are from before the wave of deep learning algorithms in face recognition~\cite{bourlai2009matching, 2011restoring, starovoitov2002three, 2000matching}.
There are also significant recent efforts involving deep learning algorithms~\cite{2018cross, docface+}. 
In this work, we consider the additional challenge that the ID document image is from early adolescence and the live image from later adolescent / early adulthood, so the face appearance may have undergone substantial natural change.
This problem context arises, for example, in applications where a young person presents his/her ID card to enroll for a government benefit and the ID card either does not contain a digital image or the application is not authorised or equipped to access such an electronically embedded image.

Our experimental results are reported using an extended version of the Chilean Young Adults (CHIYA) dataset, composed of 263,490 image pairs.
One image of a pair is an image of a national ID card and the other is a selfie-style face image of a person claiming the same identity as illustrated in Fig. \ref{fig:problem}.
The subject is as much as 9 years younger in their ID card image than they are in the live image.

Matchers evaluated for this challenging problem include COTS (commercial off-the-shelf) and GOTS (government off-the-shelf) matchers, open-source CNN matchers, including ArcFace~\cite{arcface}, and matchers proposed in recent work on the doc-face-to-selfie problem~\cite{docface+}.
Then, using the open CNN matcher that achieves the highest accuracy, we fine-tune it using triplet loss for few-shot learning.
We use triplet loss to deal with the fact that we have just two images per person, as triplet loss does not perform classification, but instead verification, thus no convergence difficulty is observed.
To deal with the heterogeneity in the two types of images, the fine-tuning selects positive and negative pairs that are the opposite type of the anchor, thus learning to separate impostors and authentic pairs in the appropriate context.
By using more than 230,000 subjects in training, with subjects having different elapsed times for their image pairs, the fine-tuning process should learn to deal with the elapsed time problem.
Thus, our fine-tuning approach is designed with features to address the issues of (i) few images per subject, (ii) matching images of two different types, and (iii) varying age difference during adolescence for different image pairs.

The contributions of this work to the ``Document Face to Selfie Face Matching Problem'' can be summarized as follow:
\begin{itemize}
    \vspace{-0.5em}
    \item Only work to study adolescence scanned identity document to selfie matching, demonstrating the difficulty of the problem.
    \vspace{-0.5em}
    \item A freely available model with a significant increase in accuracy in the CHIYA dataset; this model also performs well on a different dataset (Public-Ivs~\cite{ivs}), and surpass previous works~\cite{ivs, docface+} on ID versus selfie matching problem.
    \vspace{-0.5em}
    \item Finding of no consistent or large gender difference in accuracy between adolescents document-to-selfie face matching.
\end{itemize}

\section{Related Work}

Two streams of previous work come together in this project.
One is the doc-face-to-selfie matching problem, and the other is the problem of face matching across varying time lapse during adolesence.

\subsection{Face Matching Across Adolescence}

Adolescence refers to the period of growth and development of children, roughly the age range of 10 to 19 years, starting with puberty and ending with an adult identity and behaviour~\cite{nihadolescentsdef}.
%
%
Ricanek et al.~\cite{Ricanek2015} assembled the ``In the Wild Celebrity Children'' (ITWCC) dataset, and evaluated the accuracy of various ``hand-engineered'' (not deep learning) face matchers.
Their dataset contains images of subjects at child, preadolescence and adolescence ages.

Best-Rowden et al.~\cite{BestRowden2016} evaluated the accuracy of a COTS algorithm on the ``Newborns, Infants, and Toddlers'' (NITL) dataset, and studied the effects of age variation.
High verification accuracy was observed when comparing images that are captured in the same session and a degradation in accuracy was observed when comparing images captured in different sessions.
%
%

Deb et al.~\cite{Deb2018} investigated the accuracy of a COTS matcher, the FaceNet~\cite{facenet} open-source matcher, and a fusion of the two, on the Children Longitudinal Face (CLF) dataset.
They investigated verification and identification scenarios, for age differences of 1 year, 3 years and 7 years, and found decreased accuracy as age difference increases. 
They found that fine-tuning the FaceNet matcher based on a set of $3,294$ face images of $1,119$ children aged 3 to 18 years old improved accuracy over the default FaceNet.

Using the ITWCC dataset, Srinivas et al.~\cite{nisha_wacv2019} evaluate the performance of child face recognition using several COTS and GOTS deep learning based face matchers.
(They find that child face recognition has a much lower accuracy than adult face recognition.)
This work also suggests that there is some element of gender bias in child face recognition, as all the matchers considered had lower accuracy for females.

Srinivas et al.~\cite{nisha_cvpr2019} studied the difference in accuracy between child and adult face matching, using five top performing COTS matchers as rated in a NIST study\footnote{https://www.nist.gov/programs-projects/face-recognition-vendor-test-frvt}, two GOTS and one open-source matcher.
They found a much lower accuracy for all matchers for child faces than for adult faces.
To improve the accuracy for child faces, the authors proposed several score-level fusion techniques, which improved accuracy over individual matchers.

Michalski et al.~\cite{Michalski2018} used a dataset of $4,562,868$ face images of children captured under a controlled environment, similar to standard passport or visa images, to study the effects of age (0-17 years) and age variation (0-10 years) on the accuracy of a COTS matcher. 
The authors focus on an operational use of face recognition, i.e., the use of matching child visitors to their e-passports, and the need to assign different thresholds for matching when the subject is a child. 
The authors conclude that it is harder to match infants, toddlers, and young children, regardless of age variations. 

Points of note for this stream of work include:
\begin{itemize}
    \vspace{-0.5em}
    \item No previous analysis in juvenile phase from scanned ID documents face matching.
    \vspace{-0.5em}
    \item Only~\cite{Michalski2018} uses a private dataset that contains more people or images than the CHIYA dataset; however, \cite{Michalski2018} reports accuracy for a single COTS matcher.
    \vspace{-0.5em}
    \item Accuracy for adolescent face matching is generally expected to be lower than for adults, and lower for female than male adolescents~\cite{nisha_wacv2019}.
\end{itemize}

\subsection{Document Face to Selfie Face Matching}

Using a dataset collected in a real banking scenario, called FaceBank, Oliveira et al.~\cite{2018cross} used pre-trained CNNs to extract features to match ID cards to selfies.
Their private dataset consist of $13,501$ subjects, which had their ID cards and faces collected by Brazilian bank systems.
The authors trained and tested different classifiers using the difference between the deep features extracted from ID cards and a selfie.
The classifiers were trained to predict if the pair is authentic or impostor, and showed accuracy rates higher than $93\%$.

Zhu et al.~\cite{ivs} studied matching ID cards versus selfies in CASIA-IvS dataset.
Using a dataset of a million scale, the authors proposed a three-stage training, where in the first stage the model is trained on general in-the-wild faces for classification, in the second stage the model is fine-tuned on the CASIA-IvS dataset for verification, and finally the last stage is fine-tuned for classification on the CASIA-IvS dataset using a novel softmax approach.
The authors released a small testing set, called Public-IvS. However, this dataset is not formed by real ID cards, but rather ``simulated'' cards. Simulation achieved by selecting images with an uniform background.

DocFace+~\cite{docface+} proposes a similar approach, where the authors first trained their model on a traditional in-the-wild dataset, and then fine-tuned on their private ID card versus selfies dataset, which is composed of selfie photos and passport embedded ID photos.
The fine-tuning is performed using a pair of sibling networks, and a novel optimization method that was designed for training on shallow datasets.
The optimization method, called dynamic weight imprinting (DWI), works by updating only the weights of the classes that are present in the batch being processed, thereby helping the training to converge.
The authors show that their proposed method surpasses their base model, and the face matchers evaluated on their private dataset.
The authors also tested their model on the Public IvS, and concluded that because it is not a real ID versus selfie dataset, their fine-tuned model performed worse than its pre-trained version. 


Points to note from this stream of previous work include:
\begin{itemize}
    \vspace{-0.5em}
    \item No previous work on document-to-selfie face matching considers the adolescent age range.
    \vspace{-0.5em}
    \item Only~\cite{2018cross} works with document face images acquired by taking an image of the ID document, as distinguished from digital images stored as part of the ID document.
    \vspace{-0.5em}
\end{itemize}

\begin{figure}[t]
    \centering
    \includegraphics[width=0.7\columnwidth]{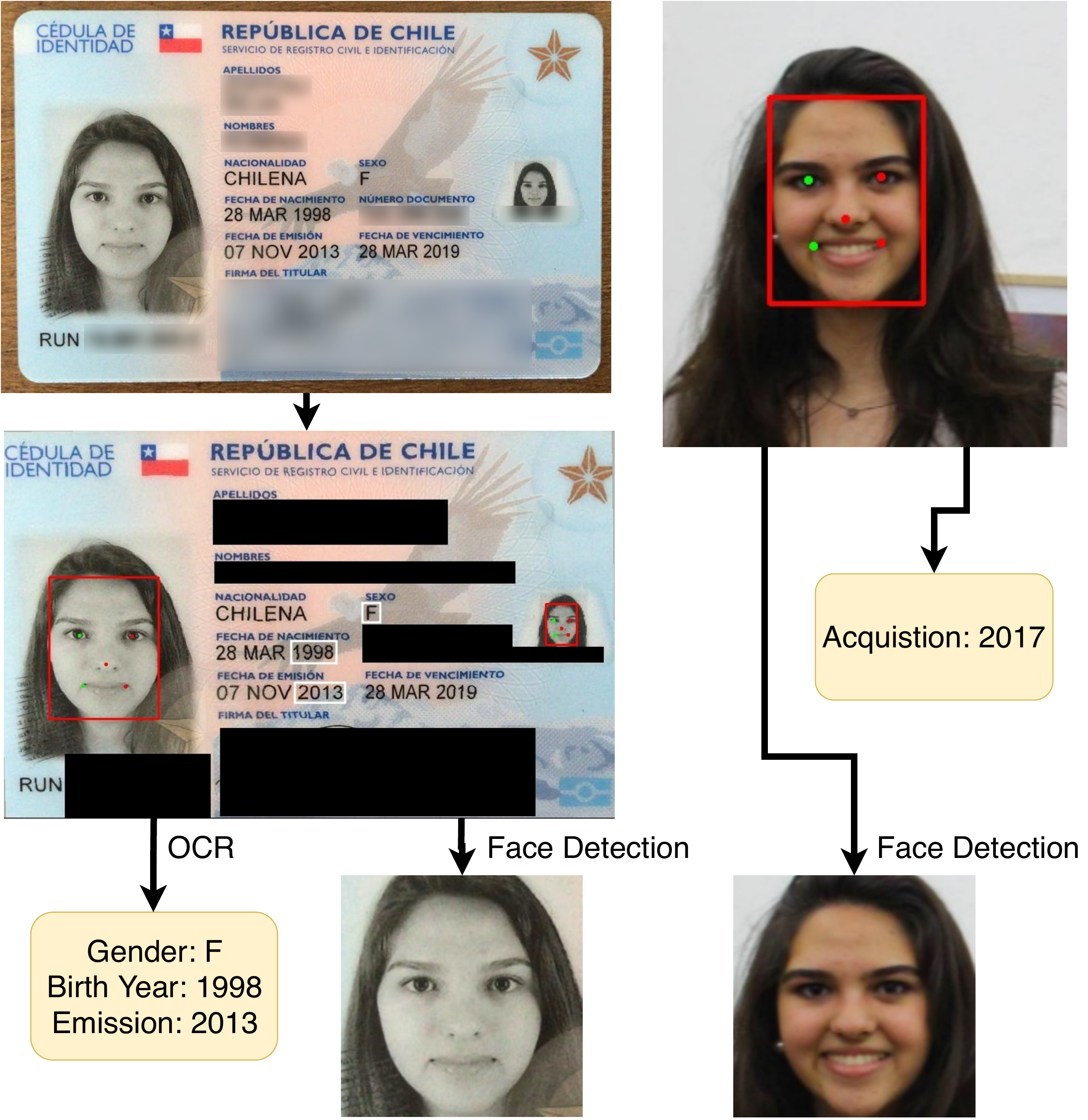}
    \caption{Overview of pre-processing: the ID cards are segmented and the confidential information removed, then face detection and OCR is applied; face detection is applied on the selfie images and acquisition date extracted.}
    \label{fig:preprocessing}
    \vspace{-1em}
\end{figure}

\section{Experimental Dataset}
\label{sec:dataset}



We collected a dataset called Chilean Young Adult (CHIYA), that contains pairs of images acquired by persons enrolling for a benefit in an operational scenario since 2017.
Each pair of images is one image of the person's Chilean national ID card, issued at an earlier age, and one selfie-style current image of the person, acquired with a contemporary mobile device (see Fig. \ref{fig:problem}).
The particular device used to acquire a pair of images vary between image pairs.
CHIYA currently has a total of $263,490$ subjects, each with one pair of images, for a total of $526,580$ images.
Further, the Chilean national ID card format is of two different formats, a ``yellow card'' and a later ``blue card'', due to a change in the card format by the government in year 2013.
The card formats are discussed in a later section.

Before face detection, pre-processing steps are performed for the ID cards.
SIFT points and homography are used to segment and warp the card to a flat view, and confidential information is redacted, e.g. name, id number, etc.
OCR\footnote{Google’s Tesseract-OCR: https://pypi.org/project/pytesseract/} is then applied to the card to obtain estimated values for gender, year of birth and year ID card was issued.
The year of birth and the year of ID card was issued can be used to estimate age of the person.

Next, we use RetinaFace~\cite{retinaface} to detect and align the ID card face and the selfie face, which are then resized to 224x224 pixels.
$20,262$ subjects that did not have a face detected in the ID card image or the selfie image or had problems with the warp transformation were removed.
Figure \ref{fig:preprocessing} shows an overview of the pre-processing steps.
The dataset after pre-processing consists of $243,228$ subjects and $486,456$ images.
For about $10,000$ of the subjects, for the ID card image, the face detection result and the OCR results for age and gender were manually verified.
From this curated subset, we created five subgroups based on their age on the ID cards and their age at selfie acquisition. 
The nomenclature of the subgroups is as follows: $iXXsYY$, where $i$ indicates the ID image, $XX$ indicates the age of the subject on the id image, $s$ indicates the selfie image, and $YY$ indicates the age of the subject recorded when the selfie image was captured. 
The number of subjects are summarized in Table \ref{tab:test_subsets}.

\begin{table}[t]
    \centering
    \caption{Testing subsets demographics.}
    \begin{tabular}{l|rrr}
        \textbf{Subset Name} & \textbf{Males} & \textbf{Females} & \multicolumn{1}{l}{\textbf{Total}} \\ \hline
        i10s1819: 10 vs 18-19& 324& 277& 631\\
        i12s1819: 12 vs 18-19& 1,182& 828& 2,010\\
        i14s1819: 14 vs 18-19& 1,545& 1,075& 2,620\\
        i16s1819: 16 vs 18-19& 919& 700& 1,619\\
        i18s1819: 18 vs 18-19& 1,396& 1,246& 2,642 
    \end{tabular}
    \label{tab:test_subsets}
    \vspace{-1em}
\end{table}

\begin{table*}[t]
    \setlength\tabcolsep{3pt}
    \small
    \centering
    \caption{Comparison of existing methods on the five evaluation sets.}
    \begin{tabular}{l|ccccc|ccccc}
        \textbf{}& \multicolumn{10}{c}{\textbf{True Acceptance Rate (\%)}} \\
        \multicolumn{1}{l|}{}& \multicolumn{5}{c|}{\textbf{False Acceptance Rate = 0.01\%}} & \multicolumn{5}{c}{\textbf{False Acceptance Rate = 0.1\%}}\\
        \multicolumn{1}{l|}{\textbf{Method}} & \textbf{i10s1819} & \textbf{i12s1819} & \textbf{i14s1819} & \textbf{i16s1819} & {\textbf{i18s1819}} & \textbf{i10s1819} & \textbf{i12s1819} & \textbf{i14s1819} & \textbf{i16s1819} & \textbf{i18s1819} \\ \hline
        
        
        COTS-1 & 0.63& 0.85& 10.19 & 12.97 & 14.8& 2.85& 4.68& 27.98 & 35.02 & 36.03 \\
        GOTS-1 & 15.21 & 24.73 & 44.2& 60.84 & {69.11} & 35.97 & 46.07 & 69.43 & 82.83 & 87.66 \\
        GOTS-2 & 28.68 & 36.02 & 64.5& 80.11 & {85.28} & 47.86 & 58.01 & 83.05 & 92.03 & 94.7\\
        VGGFace \cite{vgg-face} & 3.8 & 4.68& 13.05 & 24.95 & {32.36} & 10.78 & 13.83 & 30.15 & 45.89 & 54.39 \\
        FaceNet \cite{facenet}& 9.51& 11.59 & 30.99 & 46.02 & {54.47} & 21.71 & 27.16 & 54.43 & 72.95 & 77.63 \\
        VGGFace2 \cite{vggface2}& 12.36 & 14.93 & 34.43 & 51.2& {60.18} & 26.15 & 33.88 & 55.84 & 75.85 & 80.55 \\
        DocFace+ base model \cite{docface+} & 27.26 & 36.57 & 69.92 & 83.08 & {86.68} & 48.81 & 56.77 & 85.61 & 92.77 & 94.63 \\
        DocFace+ \cite{docface+} & 35.18 & 39& 63.02 & 75.36 & 78.99 & 53.41 & 59.3& 79.77 & 88.45 & 90.92 \\
        ArcFace \cite{arcface}& \textbf{62.92} & \textbf{76.52} & \textbf{93.02} & \textbf{95.86} & \textbf{97.31} & \textbf{80.82} & \textbf{89.45} & \textbf{97.79} & \textbf{98.7}& \textbf{98.9}\\
    \end{tabular}
    \label{tab:methods_comparison}
    \vspace{-1em}
\end{table*}

\section{Face Matchers}

We report experimental results for: one commercial (COTS-1) and two government methods (GOTS-1 and GOTS-2);
four well-known deep learning face matchers (VGGFace~\cite{vgg-face}, FaceNet~\cite{facenet}, VGGFace2~\cite{vggface2} and ArcFace~\cite{arcface}), and another
deep learning method (DocFace+ base model~\cite{docface+}) from a previous research work in this area;
one model (DocFace+~\cite{docface+}) optimized for the document-to-selfie face matching task;
two of the initial higher-accuracy models with fine-tuning on the CHIYA dataset, and finally, our proposed method.

For the COTS and GOTS matchers, the images were kept at their original size (224x224 pixels), and the matchers' own face detection was skipped.
VGGFace and VGGFace2 used the original images with 224x224 pixels.
For FaceNet, the images were resized to 160x160 pixels in keeping with its normal input image size.
Similarly, ArcFace uses an image size of 112x112 pixels.
For DocFace+ the images were resized to 112x112 pixels and then center cropped to 96x112 pixels, as the model uses rectangles of this size.
The models trained on the CHIYA dataset use the same image size as their pre-trained counterpart.
Finally, except for the COTS and GOTS matchers which generated their own similarity scores, the feature vectors were matched using cosine similarity.




\section{Few-Shot Learning with Triplet Loss}
\label{sec:fine-tuning}

\begin{figure}[t]
    \centering
    \includegraphics[width=0.7\columnwidth]{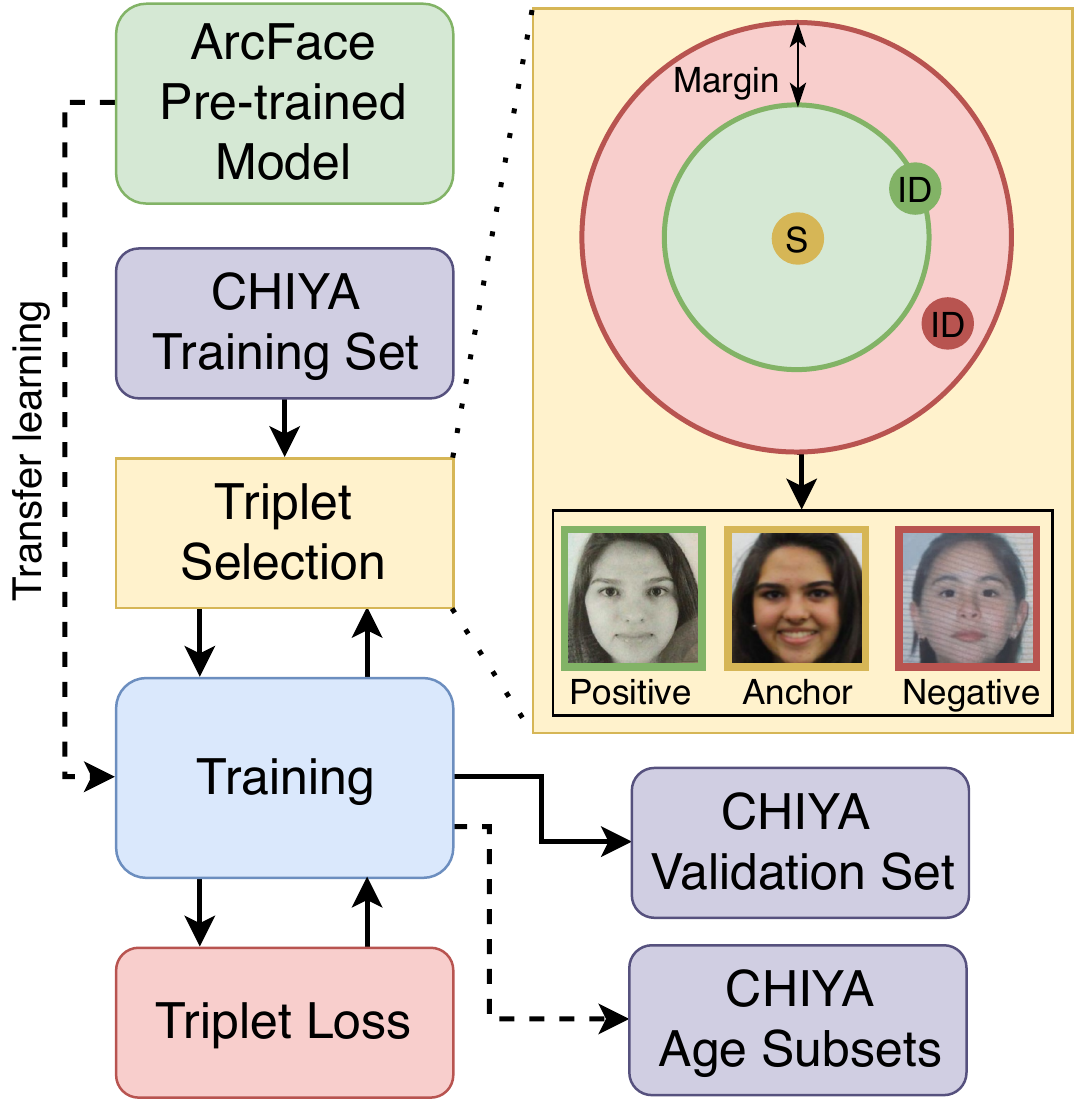}
    \caption{Overview of the proposed method. Dashed lines mean step happens only once. The CHIYA dataset has more than $260,000$ subjects, and has five curated subsets with increasing age (10 to 18) in the ID card against selfies of 18-19 years old teenagers.}
    \label{fig:overview}
    \vspace{-1.5em}
\end{figure}

Our dataset has over 200,000 subjects and just two images per subject, so training a model from scratch, or fine-tuning a model in a classification context, is not feasible.
Instead, we approach the problem as a few-shot learning using triplet loss, with transfer-learning from a model that was trained on a larger, general in-the-wild dataset.
Semi-hard triplet selection is performed as in~\cite{facenet}.
However, as we are only interested in matching between selfies and ID card images, all the triplets are selected as selfies to ID cards.
That is, if the anchor is a selfie face, the positive can only be an ID card face (as the dataset only has two images per subject), and we constrain the negative to also be an ID card face.
In this way, the network learns only from triplets that are meaningful for the task.

We select ArcFace~\cite{arcface} as the pre-trained model for fine-tuning, as it shows the best performance in our initial experiments (as shown in next section).
The pre-trained ArcFace model~\cite{insightface} used was trained with ResNet-100~\cite{resnet} on the MS1MV2 dataset, which is a cleaned version of the MS1M dataset~\cite{ms1_celeb}.

As fine-tuning data, we use $231,008$ subjects (with two images each) that are disjoint from the subjects in the five testing sets listed in Table~\ref{tab:test_subsets}.
The training data is split into $90\%$ training and $10\%$ validation.
However, as many pairs in the validation set will be easy, we do not use the entire $10\%$ validation.
Instead, we split the validation into ten subject disjoint folders, which are then matched, creating ten impostor and ten authentic distributions.
Then, for each authentic distribution, we select the $300$ subjects with the lowest-similarity image pairs (hard positive pairs).
After that, for each subject selected, we select the highest similarity impostor among the selected subjects, to create an additional $300$ impostor pairs (hard negative pairs).
The final validation set is composed of ten folders with $600$ pairs each ($300$ authentic and $300$ impostor), totalling $6,000$ comparisons (which follows the size of commonly-used validation sets~\cite{agedb, lfw}).
Finally, we measure the accuracy on the validation set as the true acceptance rate (TAR) at a false acceptance rate (FAR) of $0.1\%$, which is used to select the fine-tuning best weights.

We fine-tuned the ArcFace model using a learning rate of $0.005$, mini-batch size of $240$, and a margin of $0.3.$
Stochastic Gradient Decent (SGD) was used with a momentum of $0.9$.\footnote{No parameter search was conducted on the fine-tuning.}
We evaluate the model every $200$ iterations, and the best weights were selected at iteration $2,400$.
To make the distinction between models, we refer to our fine-tuned ArcFace model as ``AIM-CHIYA'' (ArcFace Id Matching on CHIYA).
Figure \ref{fig:overview} shows an overview of the proposed method.

\section{Experiments and Results}

This section is organized as follows: 
Section \ref{sec:methods_comparsion} discusses the accuracy of COTS, GOTS, and versions of open-source matchers not fine-tuned on the CHIYA dataset.
Section \ref{sec:arcfacedoc_results} compares AIM-CHIYA results to the base ArcFace model.
Section \ref{sec:docface_comparison} compares AIM-CHIYA to DocFace+ and DocFaec+ base model fine-tuned on the CHIYA dataset.
Section \ref{sec:gender_analysis} analyzes the accuracy difference by gender (recall that previous research suggested~\cite{nisha_wacv2019} that females adolescents have lower face matching accuracy than males).
Finally, Section \ref{sec:card_analysis} compares accuracy for the two formats of Chilean national ID cards.
If not stated otherwise, all TAR results in this section are with a FAR of $0.01\%$.

\subsection{Accuracy Evaluation of Existing Methods}
\label{sec:methods_comparsion}

Table \ref{tab:methods_comparison} shows each matcher's accuracy on the five age-span test sets defined in Table \ref{tab:test_subsets}. 
We can see that the problem increases in difficulty as the age span increase between ID card and selfie increases.
This agrees with the general expectation for ``template aging'' in face recognition (and other biometric modalities).
COTS-1 had the poorest accuracy, showing an average TAR of $7.89\%$ across the testing subsets.
There is significant accuracy difference between GOTS-1 with $42.82\%$, and GOTS-2 with $58.91\%$.
The highest accuracy matchers were the DocFace+ base model, DocFace+ and ArcFace.

\begin{figure}[t]
    \centering
    \begin{subfigure}[b]{1\linewidth}
        \centering
        \includegraphics[width=1\columnwidth]{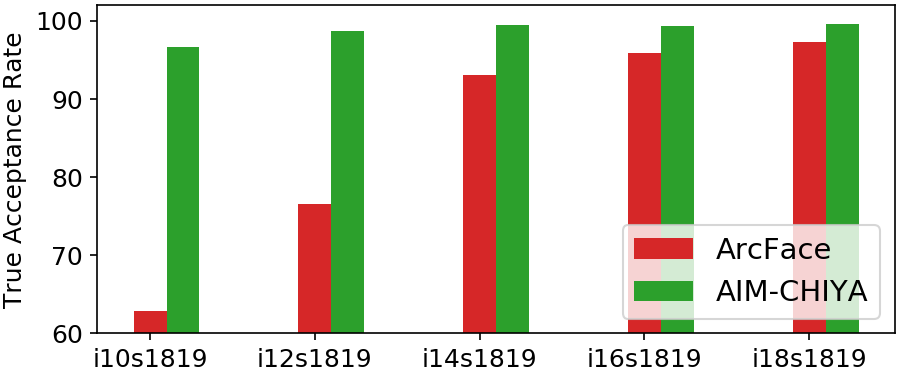}
        \caption{False Acceptance Rate = 0.01\%}
    \end{subfigure}
    \begin{subfigure}[b]{1\linewidth}
        \centering
        \includegraphics[width=1\columnwidth]{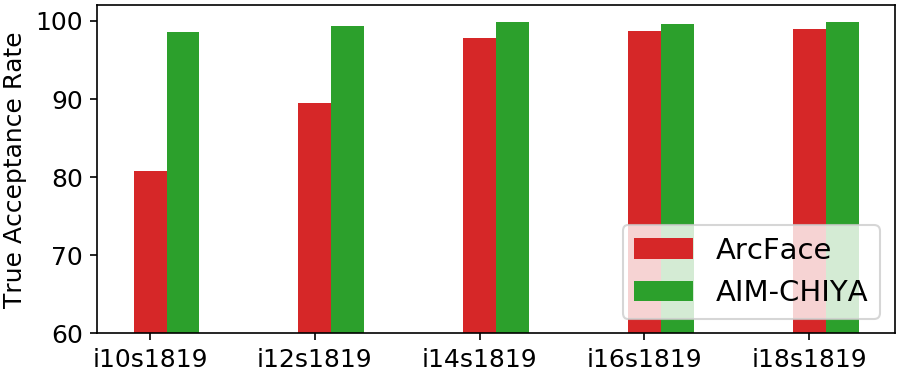}
        \caption{False Acceptance Rate = 0.1\%}
        \vspace{-0.5em}
    \end{subfigure}
    \caption{ArcFace vs. AIM-CHIYA (ours) on the five testing sets.}
    \label{fig:arcface_arcfacedoc}
    \vspace{-1em}
\end{figure}

The accuracy among the deep learning matchers varies greatly, from an average TAR of $15.77\%$ (VGGFace) to $85.13\%$ (ArcFace).
DocFace+, which is the only matcher specialized for the document-to-selfie matching task, performs worse than ArcFace and the DocFace+ base model.
DocFace+ having lower accuracy than the DocFace+ base model is contrary to what might be expected.
However, we speculate plausible reasons: first, DocFace+ was fine-tuned on an ID card dataset of Asian faces, which may show ``other race effect''~\cite{phillips2011other} when testing on Chilean faces; second, the ID photos were read from the ID document chip (better quality) in the DocFace+ dataset, but are from a photograph of the card (lower quality) in CHIYA.

ArcFace shows the best overall performance across all subsets with a significant margin over the second highest-accuracy matcher, and is especially good on the hardest subsets.
However, even with the highes-accuracy matcher, it is difficult is to match face images of adolescents with a substantial time lapse between ID cards and selfies.
The difference between the hardest test set (is10s1819) and the easiest (i18s1819) is $34.39\%$.

\subsection{Proposed Method (AIM-CHIYA) Accuracy}
\label{sec:arcfacedoc_results}

Figure \ref{fig:arcface_arcfacedoc} shows the comparison between ArcFace and ArcFace fine-tuned, or ``AIM-CHIYA'', as described in Section \ref{sec:fine-tuning}.
All five time-lapse test groups increased their accuracy, and the hardest two subsets show the biggest increases.
The subset i10s1819 went from a TAR of $62.92\%$ to $96.67\%$, and the subset i12s1819 from $76.52\%$ to $98.71\%$.
The gap in accuracy between the hardest and easiest subset was reduced to just $2.95\%$.
This demonstrates that our proposed method is successful in increasing matching accuracy to acceptable operational levels, with similar accuracy across all age subsets.

\subsection{Comparison with DocFace+ Fine-tuned Models}
\label{sec:docface_comparison}

\begin{figure}[t]
    \centering
    \begin{subfigure}[b]{1\linewidth}
        \centering
        \includegraphics[width=1\columnwidth]{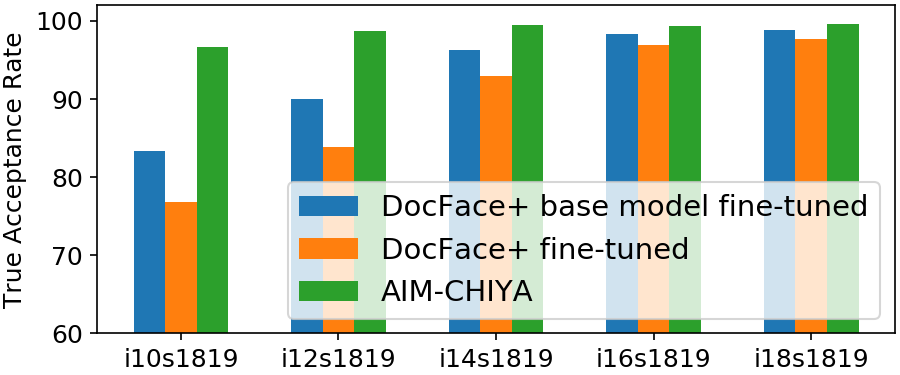}
        \caption{False Acceptance Rate = 0.01\%}
    \end{subfigure}
    \begin{subfigure}[b]{1\linewidth}
        \centering
        \includegraphics[width=1\columnwidth]{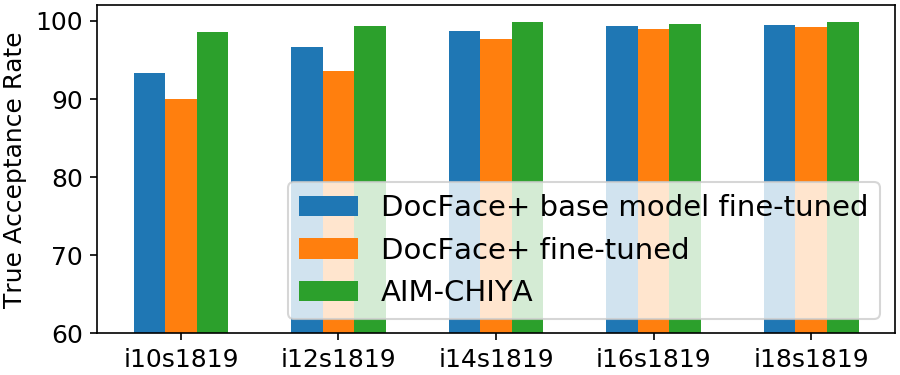}
        \caption{False Acceptance Rate = 0.1\%}
    \end{subfigure}
    \caption{Comparison of both DocFace+ models fine-tuned on the CHIYA dataset against AIM-CHIYA (ours).}
    \label{fig:docface_arcfacedoc}
    \vspace{-1em}
\end{figure}

To attempt a fair comparison between AIM-CHIYA and DocFace+~\cite{docface+}, we fine-tuned both DocFace+ and the DocFace+ base model on the CHIYA dataset using the authors' implementation~\cite{docface+_github}.
We used the authors' default parameters, and trained for 22,400 iterations using the same training and validation sets as for AIM-CHIYA.
We observed some instability in fine-tuning the models, with some training attempts not converging at all (accuracy always around $0.001\%$) or showing lower accuracy than expected.
The problem seems to go away with a restart of the training, which suggests that the DocFace+ fine-tuning is overly sensitive to the random order of data being selected.

The results of both the DocFace+ base model fine-tuned and DocFace+ fine-tuned are shown on Figure \ref{fig:docface_arcfacedoc}.
First, the DocFace+ base model fine-tuned shows better performance than the DocFace+ fine-tuned, as would be expected given the results of both pre-trained versions of the models.
Comparing to AIM-CHIYA, both DocFace+ fine-tunings show lower accuracy, with decreasing difference as the subset decreases in difficulty.
The largest difference between DocFace+ base model fine-tuned and AIM-CHIYA is $13.31\%$, and the smallest is $0.65\%$, for the hardest (i10s1819) and easiest (i18s1819) subsets, respectively.

\subsection{Accuracy Difference By Gender}
\label{sec:gender_analysis}

\begin{figure}[t]
    \centering
    \begin{subfigure}[b]{1\linewidth}
        \centering
        \includegraphics[width=1\columnwidth]{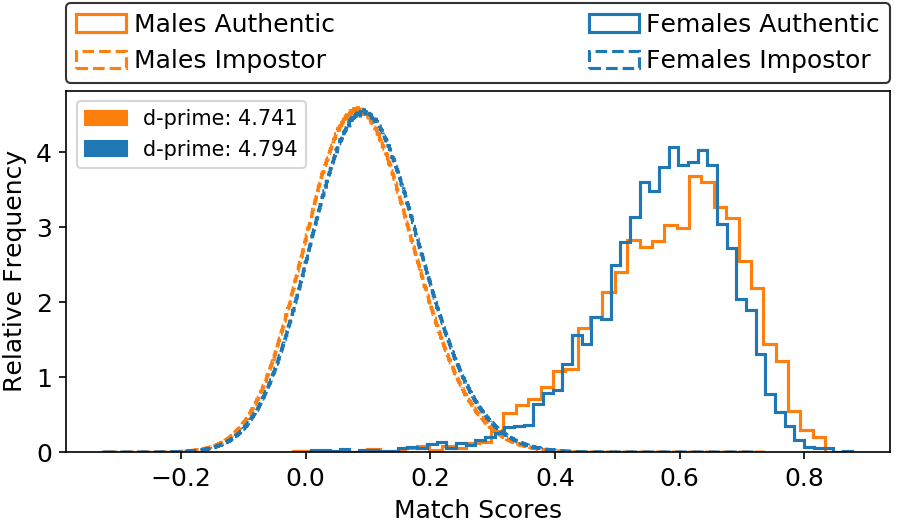}
        \caption{ArcFace}
    \end{subfigure}
    \hfill
    \begin{subfigure}[b]{1\linewidth}
        \centering
        \includegraphics[width=1\columnwidth]{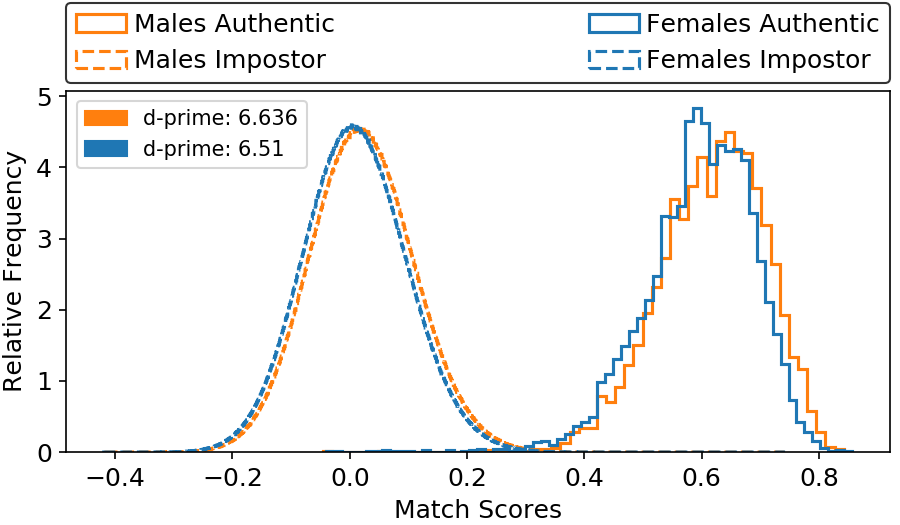}
        \caption{AIM-CHIYA}
    \end{subfigure}
    \caption{ArcFace and AIM-CHIYA (ours) males and females match score distribution.}
    \label{fig:distributions}
    \vspace{-1em}
\end{figure}
\begin{figure}[t]
    \centering
    \begin{subfigure}[b]{1\linewidth}
        \centering
        \includegraphics[width=1\columnwidth]{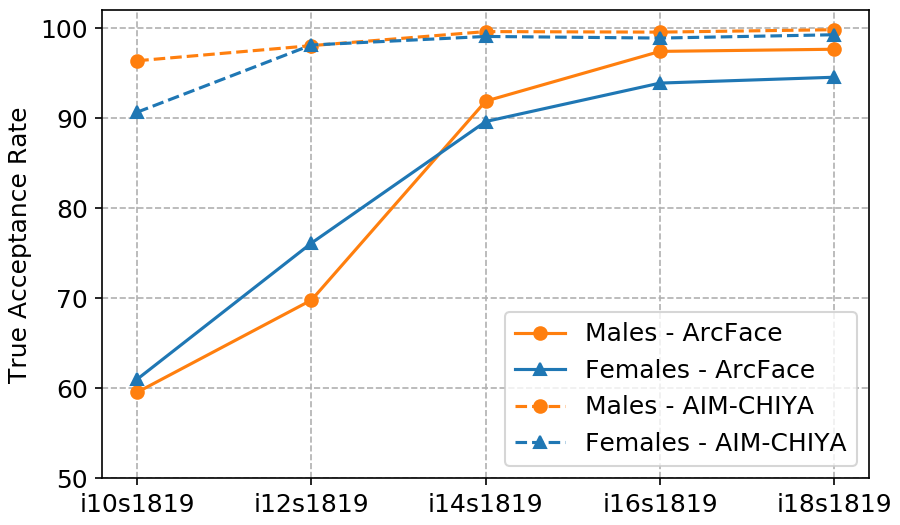}
        \caption{False Acceptance Rate = 0.01\%}
    \end{subfigure}
    \begin{subfigure}[b]{1\linewidth}
        \centering
        \includegraphics[width=1\columnwidth]{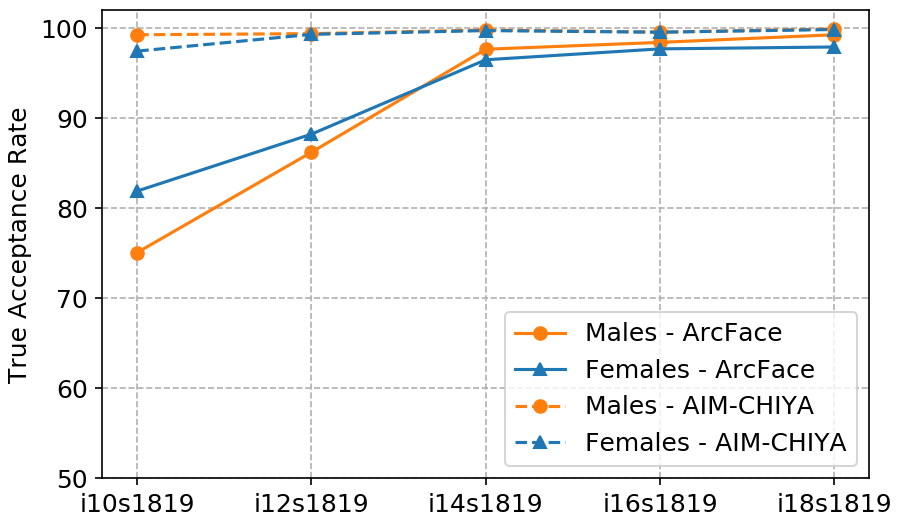}
        \caption{False Acceptance Rate = 0.1\%}
    \end{subfigure}
    \caption{ArcFace and AIM-CHIYA (ours) gender accuracy.}
    \label{fig:arcface_males_females}
    \vspace{-1em}
\end{figure}

Using the best general face matcher, ArcFace, and our proposed method, AIM-CHIYA, we analyze the performance across males and females.
Combining all the groups together, the authentic and impostor match scores are shown in Figure \ref{fig:distributions}.
In the base ArcFace distributions, the female impostor curve is shifted to higher (worse) values than the males.
However, after fine-tuning, the pattern inverts, and males have an impostor curve shifted to greater-similarity values.
Even though the authentic distributions are not completely Gaussian, the d-prime for males and females are very close in both ArcFace and AIM-CHIYA.
The TAR of ArcFace is $87.49\%$ (same) for both males and females, and at a higher FAR (0.1\%), the TAR is only slightly different, $94.26\%$ and $94.55\%$ for males and females, respectively.
For AIM-CHIYA, the TAR is $99.35\%$ and $98.78\%$ for males and females respectively, showing that males had their accuracy increase a little more than females.

Figure \ref{fig:arcface_males_females} shows the accuracy difference between males and females across subsets.
For ArcFace, females show higher accuracy on the hardest two subsets (i10s1819 and i12s1819).
On the other hand, AIM-CHIYA shows better performance for males on the hardest subset (i10s1819), but the accuracy is much more similar for the other four subgroups.
This demonstrates that fine-tuning ArcFace is able to improve the accuracy of both males and females by a similar amount.

Overall, subjectively, there does not appear to be any large consistent difference with gender. 
The small observed differences flip between matchers and for different age ranges.

\subsection{Comparison Between ID Card Formats}
\label{sec:card_analysis}

\begin{figure}[t]
    \centering
    \begin{subfigure}[b]{0.28\linewidth}
        \centering
        \includegraphics[width=\linewidth]{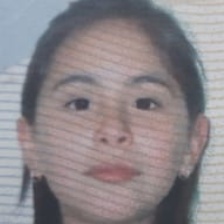}
        \caption{Yellow Card}
        \vspace{-0.5em}
    \end{subfigure}
    \hspace{1em}
    \begin{subfigure}[b]{0.28\linewidth}
        \centering
        \includegraphics[width=\linewidth]{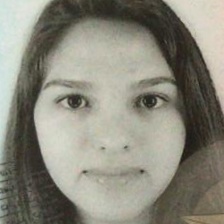}
        \caption{Blue Card}
        \vspace{-0.5em}
    \end{subfigure}
  \caption{Yellow and blue card sample aligned faces of different subjects.}
  \label{fig:card_examples}
\end{figure}

As mentioned in Section \ref{sec:dataset}, the CHIYA dataset contains two formats of ID cards.
The yellow ID card has lower quality face image, as it has watermark stripes overlayed.
The blue ID card does not contain watermark, but the face image is grayscale rather than color. 
Figure \ref{fig:card_examples} shows an example of each card format.

Table \ref{tab:card_types} shows the number of subjects with each format of ID cards across subsets.
There is not a single subset that has enough subjects with both formats of ID card for matching and computing accuracy values.
Instead, we analyze the mean average authentic and impostor scores for both ID cards.
This was performed on the subgroups i12s1819 and i14s1819, as they are the only two groups with enough images of both formats of cards.
In the subset i12s1819, we randomly selected ten times $74$ subjects that have an yellow ID card, and in the subset i14s1819, we randomly selected ten times $138$ subjects with blue ID cards, so to match the number of the opposite format of card in that subset.

\begin{figure*}[t]
    \centering
    \begin{subfigure}[b]{0.47\linewidth}
        \centering
        \begin{subfigure}[b]{0.48\linewidth}
            \centering
            \includegraphics[width=1\columnwidth]{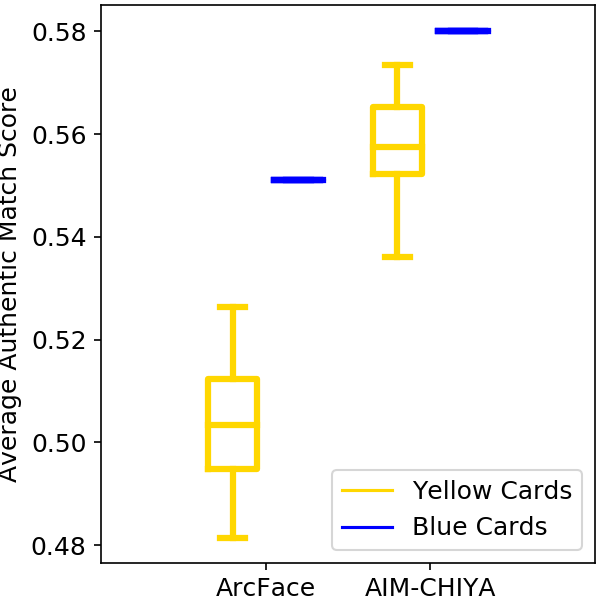}
        \end{subfigure}
        \hfill
        \begin{subfigure}[b]{0.48\linewidth}
            \centering
            \includegraphics[width=1\columnwidth]{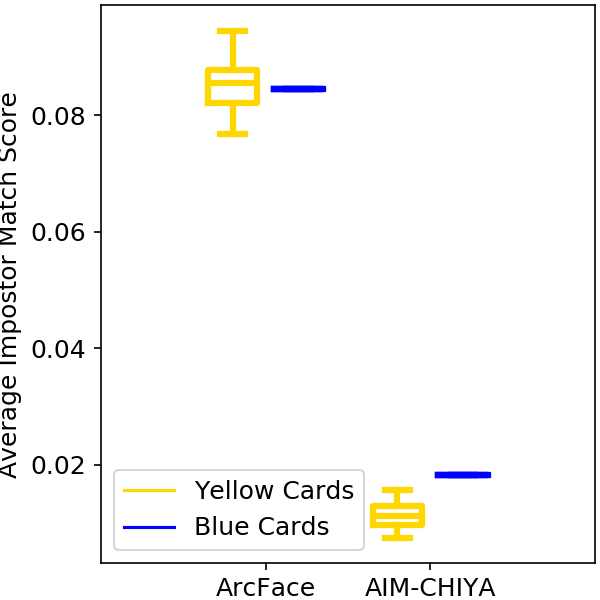}
        \end{subfigure}
        \caption{i12s1819}
        \vspace{-0.5em}
    \end{subfigure}
    \hfill
    \begin{subfigure}[b]{0.47\linewidth}
        \centering
        \begin{subfigure}[b]{0.48\linewidth}
            \centering
            \includegraphics[width=1\columnwidth]{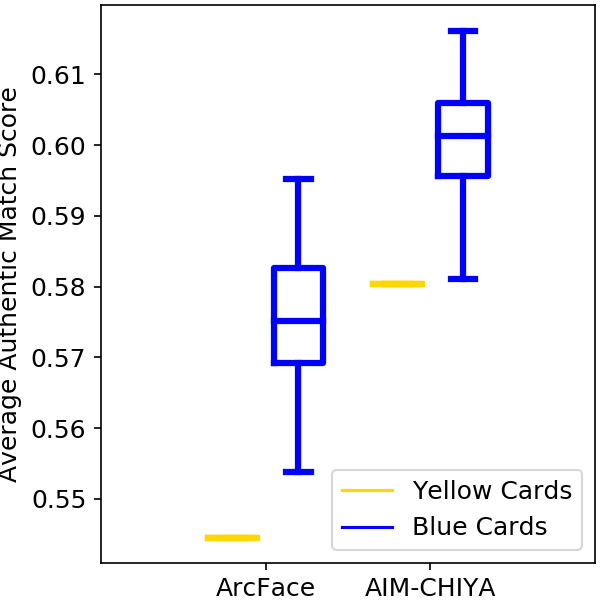}
        \end{subfigure}
        \hfill
        \begin{subfigure}[b]{0.48\linewidth}
            \centering
            \includegraphics[width=1\columnwidth]{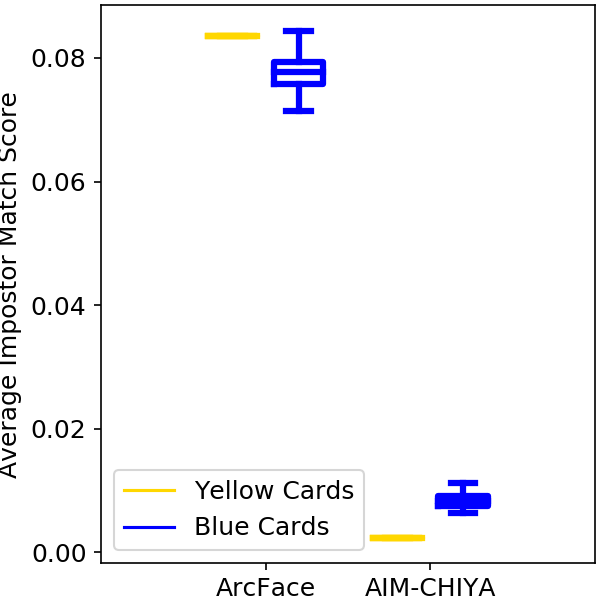}
        \end{subfigure}
        \caption{i14s1819}
        \vspace{-0.5em}
    \end{subfigure}
    \caption{ArcFace and AIM-CHIYA (ours) comparison between card types on two subgroups.}
    \label{fig:card_types}
    \vspace{-1em}
\end{figure*}
\begin{table}[t]
    \centering
    \small
    \setlength\tabcolsep{2.5pt}
    \caption{Testing subsets with number of each card format.}
    \begin{tabular}{l|ccccc}
        \textbf{Card Type} & \textbf{i10s1819} & \textbf{i12s1819} & \textbf{i14s1819} & \textbf{i16s1819} & {\textbf{i18s1819}} \\ \hline
        Yellow & 630 & 1936 & 138 & 2 & 3\\
        Blue & 1 & 74 & 2482 & 1617 & 2639\\
    \end{tabular}
    \label{tab:card_types}
    \vspace{-1em}
\end{table}

Figure \ref{fig:card_types} shows the average authentic and impostor match score for both formats of ID cards using ArcFace and AIM-CHIYA.
For both groups and matchers, the blue ID cards show a higher average authentic score.
However, the pattern is different for the average impostor matching score across matchers.
For ArcFace, the blue ID cards are lower than the mean impostor score on the i12s1819 subgroup, and the yellow ID cards show to be on the higher percentile of the blue ID cards on the i14s1819 subgroup.
After fine-tuning, AIM-CHIYA shows much lower scores for both formats of ID cards, with the yellow format being even lower than the blue card.
As yellow cards have a higher impostor score and a lower genuine score, we conclude that the number of yellow cards in the hardest subsets is another factor making that subset harder.
The fine-tuning of AIM-CHIYA shows that it was able to approximate the yellow ID cards authentic average scores to the blue cards, but more importantly, it made the impostor scores of yellow cards lower than the blue ones.

\section{Experiments on the Public-IvS Dataset}

To compare to previous works~\cite{ivs, docface+} and show that our model could generalize to a different dataset, we use the Public-IvS dataset~\cite{ivs}, as it is the only generally available ID card versus selfies dataset.
The Public-IvS dataset is composed of one or more ID cards per subject, as well as one or more selfie.
The dataset is also mainly Asian adult males, which is different from CHIYA on both ethnicity and age range.
And, actually, the ID card images are simulated using images with uniform background.
Not mentioned in previous works~\cite{ivs, docface+}, this dataset contains many images with more than one face.
After the face detection and alignment performed using RetinaFace~\cite{retinaface}, we manually selected the correct face when there was more than one in an image, so that all matchers tested use the same pairs.
We notice repeated identities and mislabeled images, however, we do not fix these problems, to allow a more direct comparison with previous methods~\cite{docface+, ivs}.
Finally, from the $5,503$ images available, we were able to detect a face in $5,502$.
We will make this curated version of the Public-IvS dataset available for future research comparison.

\begin{table}[t]
    \centering
    \small
    \caption{Results on the Public-IvS dataset. The results from~\cite{ivs} were reported in their paper. General matchers are show in the first two rows, and task specific matchers in the last five rows.}
    \setlength\tabcolsep{4pt}
    \begin{tabular}{l|c|c|c}
        \textbf{}& \multicolumn{3}{c}{\textbf{True Acceptance Rate (\%)}}\\
        \textbf{Method}                & \textbf{\begin{tabular}[c]{@{}c@{}}FAR\\ 0.001\%\end{tabular}} & \textbf{\begin{tabular}[c]{@{}c@{}}FAR\\ 0.01\%\end{tabular}} & \textbf{\begin{tabular}[c]{@{}c@{}}FAR\\ 0.1\%\end{tabular}} \\ \hline
        DocFace+ base model \cite{docface+} & 93.65& 97.58 & 98.83\\
        ArcFace \cite{arcface} & \textbf{99.12} & \textbf{99.28} & \textbf{99.38}\\
        \hline
        Zhu et al. \cite{ivs} & 93.62& 97.21 & 98.83\\
        DocFace+ \cite{docface+}& 90.31& 95.13 & 98.34\\
        DocFace+ base model fine-tuned & 58.8 & 71.58 & 84.47\\
        DocFace+ fine-tuned & 47.67& 63.34 & 78.09\\
        AIM-CHIYA (ours) & \textbf{98.23} & \textbf{98.81} & \textbf{99.1}
    \end{tabular}
    \label{tab:public_ivs}
    \vspace{-1em}
\end{table}

Table \ref{tab:public_ivs} shows the accuracy of models designed for the task and their pre-trained counterpart.
The accuracy between Zhu et al.~\cite{ivs} and DocFace+~\cite{docface+} base model are almost identical.
DocFace+ shows a lower performance than its pre-trained version, DocFace+ base model.
For both DocFace+ models fine-tuned on the CHIYA dataset (DocFace+ base model fine-tuned and DocFace+ fine-tuned), there is a severe degradation in accuracy, with the fine-tuned version of the DocFace+ base model TAR@FAR=0.001\% dropping from 93.65\% to 58.8\%.
A similar pattern occurs for ArcFace and AIM-CHIYA, where the fine-tuned model has worse accuracy than the pre-trained version; however, the difference in accuracy is much smaller.

The drop in accuracy between fine-tuned models and pre-trained versions is assumed to be related to the fact that the Public-IvS dataset have ``simulated'' ID cards instead of real ones.
The lower accuracy of the DocFace+ models fine-tuned on the CHIYA dataset is related to the differences in ethnicity in the CHIYA and Public-IvS dataset.
This drop in accuracy is not observed with AIM-CHIYA, which even though performing lower than its pre-trained version, was able to perform well in a completely different dataset than used in its fine-tuning.

\section{Conclusion}
This work studied the problem of matching ID cards to selfies across age differences in adolescence.
Our results show that existing methods perform poorly on the task, especially when there is a large age difference (8 - 9 years) between the ID card and selfie images.
Our proposed method AIM-CHIYA is effective in improving the accuracy of the best method (ArcFace), increasing the average TAR across groups from $85.13\%$ to $98.76\%$ at a FAR of $0.01\%$.
Also, our method improves over the accuracy of the fine-tuned state-of-the-art DocFace+ model~\cite{docface+}.

Our analysis suggests that there is not a general significant difference in accuracy between males and females, as both have a similar accuracy.

Of the two types of card format in the CHIYA dataset, results show that the yellow card results in lower accuracy.
We also find that yellow cards are are a larger fraction of the hardest subgroup.
AIM-CHIYA was able to improve the accuracy on both formats, resulting in much more similar accuracy across groups, reducing the difference of TAR@FAR=$0.01\%$ between the best (i18s1819) and the worst (i10s1819) subgroups from $34.39\%$ to $2.95\%$.

Finally, we show that our proposed method AIM-CHIYA performs well on a completely different dataset, with different ``simulated'' ID cards, different ethnicity, and different subjects age (adult people), surpassing the previous ID versus selfies methods by a significant margin.

{\small
\bibliographystyle{ieee}
\bibliography{refs}
}

\end{document}